\documentclass[letter, conference, final]{IEEEtran}

\makeatletter
\def\bstctlcite{\@ifnextchar[{\@bstctlcite}{\@bstctlcite[@auxout]}}
\def\@bstctlcite[#1]#2{\@bsphack
  \@for\@citeb:=#2\do{%
    \edef\@citeb{\expandafter\@firstofone\@citeb}%
    \if@filesw\immediate\write\csname #1\endcsname{\string\citation{\@citeb}}\fi}%
  \@esphack}
\makeatother

\usepackage[vlined, ruled, linesnumbered, commentsnumbered]{algorithm2e}
\usepackage{amsmath}
\usepackage{cite}
\usepackage{color}
\usepackage{xcolor}
\definecolor{chred}{rgb}{0.8,0,0}
\definecolor{chgray}{rgb}{0.5,0.5,0.5}
\usepackage{graphicx}
\usepackage{caption}
\usepackage{subcaption}
\usepackage{dblfloatfix}
\usepackage{eqlist}
\usepackage{txfonts}
\usepackage{url}
\usepackage{footmisc}
\usepackage{booktabs}
\usepackage{balance}

\usepackage{multirow}
\usepackage[para,online,flushleft]{threeparttable}
\usepackage{geometry}
\begin{document}
\newgeometry{top=25.4mm, bottom=19.1mm, left=19.1mm, right = 19.1mm} 
\IEEEoverridecommandlockouts

\title{\LARGE \bf
Regrasp Planning Considering Bipedal Stability Constraints}

\author{Daniel S\'anchez$^{1}$, Weiwei Wan$^{1,2,*}$, Kensuke Harada$^{1,2}$,
Fumio Kanehiro$^{2}$ 
\thanks{$^{1}$Graduate School of Engineering Science, Osaka University, Japan.
$^{2}$National Inst. of AIST. *Correspondance author: Weiwei Wan {\tt\small
wan@hlab.sys.es.osaka-u.ac.jp}}%
}

\maketitle
\thispagestyle{empty}
\pagestyle{empty}

\begin{abstract}

This paper presents a Center of Mass (CoM) based manipulation and regrasp
planner that implements stability constraints to preserve the robot balance. The
planner provides a graph of IK-feasible, collision-free and stable motion
sequences, constructed using an energy based motion planning algorithm. It
assures that the assembly motions are stable and prevent the robot from falling
while performing dexterous tasks in different situations. Furthermore, the
constraints are also used to perform an RRT-inspired task-related stability
estimation in several simulations. The estimation can be used to select
between single-arm and dual-arm regrasping configurations to achieve more
stability and robustness for a given manipulation task. To validate the planner
and the task-related stability estimations, several tests are performed in
simulations and real-world experiments involving the HRP5P humanoid robot, the
5th generation of the HRP robot family. The experiment results suggest that the
planner and the task-related stability estimation provide robust behavior for
the humanoid robot while performing regrasp tasks.
\end{abstract}

\section{Introduction} 

In this paper, we present regrasp planner considering the balances of humanoid
robots. It eliminates unstable poses for the robots during manipulation, and enables the robot to perform
dexterous tasks with one or two arms, avoiding collisions and falling. Both simulations and real-world tests are
performed to validate the utility of this solution. Furthermore, the
planner is used to compare different robot stances for complicated regrasp tasks.

Regrasp planning involves several important considerations such as the initial
and goal poses of a manipulated object, the different grasping poses available
to the robot in order to handle the object, and the intermediate poses of the
robot for achieving the object's goal pose. Further
considerations and constraints are also implemented so the regrasp task can be
performed correctly, such as collision constraints that
forbid the robot from performing movements that may make it collide
with the environment or itself.
One often overlooked consideration for the successful regrasp planning and
manipulation of an object is the balance of the robot-object system, which
proves to be especially important for unfixed robotic platforms such as humanoid
robots and wheeled robot. These robots might fall out of balance while
performing these operations with heavy or asymmetrical objects.

To create a planner that considers the robot balance, this paper
includes CoM-based constraints in a regrasp planner to enhance the robustness
and stability of the robot poses for the robot-object system. By taking into account
the CoM of the system and selecting a minimum threshold distance between the
robot-object system CoM and its support polygon, unstable poses are discarded to preserve the balance
of the robot. The CoM constraints can be also used for
choosing a robot stance that maximizes the amount of stable and
collision-free poses for the regrasp task. The use of this CoM based planner
is useful for humanoid manipulation under a fixed-leg stance such as the robot
standing on one leg.

The CoM constraints in our regrasp algorithms for single-arm and dual-arm
regrasp\cite{wan2015reorientating}\cite{wan2016developing} were tested by
performing several simulations and real-world tests using the HRP5P robot, the
5th generation of HRP humanoid robots\cite{inouehrp}. For the simulations,
several regrasp tasks were performed in order to compare the change of the
robot-object system CoM movement with different thresholds. A random-biased
sampling and evaluation of robot states, inspired by RRT\cite{LaValle1998} is
also performed for single-arm and dual-arm regrasping.
The results of these
experiments provide a means to measure and compare how stable single-arm regrasp
and dual-arm regrasp tasks are performed in different circumstances, such as
different robot stances or varying object mass and CoM. These comparisons can
then be used to select a motion that provides the most stable robot posture. For
real-world tests, the planner is implemented in the HRP5P humanoid robot. The
robot is tasked to reorient an electric drill: The drill pose and position have
to be changed by using dual-arm regrasp for which our planner
identifies and eliminates several unstable poses. The robot is able to
successfully complete the given tasks while avoiding poses that place its CoM
too close to the edge of the robot support polygon, assuring a robust motion.

\section{Related Work and Contributions}

For several years, regrasp planning has been the object of attention for
different studies. When an object's pick-up grasp is incompatible with its
put-down grasp, regrasp can be performed to achieve the goal pose
\cite{stoeter1999planning}, which is often the solution for many tasks
in robotic applications.
Some early work includes the regrasp planning of robotic manipulators such as
\cite{stoeter1999planning},\cite{lozano1992handey} and
\cite{simeon2004manipulation}, but those solutions only considered one-armed
robots with a fixed position, therefore they lack specific considerations that
must be taken into account for humanoid robots such as HRP3
\cite{kaneko2008humanoid}, WALK-MAN \cite{tsagarakis2017walk} and TALOS
\cite{stasse2017talos}.

Recently, several manipulation and regrasp planners have been proposed for
dual-armed robots with a focus on efficient and collision free object
manipulation. In \cite{wan2016developing}, comparison of different
algorithms for single-arm and dual-arm regrasp were performed in
different scenarios. In \cite{chen2017manipulation} a planner for a robot to keep an
object stable under a sequence of external forces was proposed. A CoM-based
grasp pose adaptation method for picking up objects with one arm was introduced
in \cite{kanoulas2018center}. It used 3D perception and force/torque
feedback to reduce the load of joints during manipulation. In
\cite{sundaralingam2018geometric}, a planner for autonomous in-hand manipulation
using finger gaiting was presented. Also, some researchers have worked on manipulation 
planning on constraints such as in \cite{berenson2009manipulation}.
Some optimization-base motion planning 
algorithms are summarized in \cite{schulman2014}.
These work offered several solutions for regrasp planning tasks, but they failed
to consider the balance of robot-object system. In this regard, our planner
presents a CoM-based solution that implements stability constraints during
regrasp planning.

On the other hand, several studies have been done to balance walking robots. In
\cite{garcia2002classification}, for example, a series of static and dynamic
stability criteria were evaluated for different environments. 
An early work
\cite{orin1976interactive} presented the Center of Pressure Method (CPM) and
declared that a robot is dynamically stable if the projection of its CoM along the direction
of the resultant force acting on the CoM is inside its support polygon.
Since then, more refined methods for posture and stability control have been
proposed and studied: In \cite{nava2016stability} a stability analysis and
momentum based control architecture that avoided instabilities at the
zero-dynamics level was introduced. In \cite{ott2016good} an experimental
comparison between a fully model-based control approach and a biologically
inspired approach derived from human observations was shown. A whole-body
control for balancing and pose stabilization using optimization of contact
forces and Model Predictive Control (MPC) was presented in
\cite{henze2014posture}.
A stability strategy that used the robot CoM height in its control law was
proposed in \cite{koolen2016balance}.  In \cite{Atkeson2015}, a comprehensive study of Team
WPI-CMU's approach to the DARPA Robotics Challenge (DRC), was shown, it focused
on the team's strategy to avoid failures and prevent the robot from falling.
Finally, in \cite{brunner2015design} an
iterative contact point estimation method for estimating the stability of
actively reconfigurable robots was presented.

The aforementioned control methods suggested general solutions for robot walking
and static balancing, but they did not address complicated tasks such as
object manipulation. For this reason, researchers begin to study stability-based
manipulation. In \cite{harada2005humanoid} for example, a method was proposed
for achieving balance during transporting heavy objects. Also, in
\cite{stilman2010golem} a humanoid robot designed for dynamic manipulation was
presented. In \cite{teja2015optimal}, a preliminary planner that considered the
robot CoM during manipulation was proposed, without taking into account object
properties. In \cite{ramirez2016motion} a strategy for a humanoid robot to pull
a fire hose and reach a desired goal position was discussed. These work did
not take into account more dexterous tasks such as reorienting objects and
handover, which leads to drastic change in the robot-object system CoM.
Performing these more dexterous tasks can prove to be dangerous when the
mass of the object is relatively high or when the robot is in sub-optimal
stances for balance. 

Our planner addresses the aforementioned concerns by implementing CoM-based
constraints for the robot-object system, assuring that the robot does not incur in unstable poses, increasing the
static stability of the system by keeping its CoM within a minimum distance from
the boundary of its support polygon.

The CoM constrained planner is an enhanced version of our previous planner
\cite{wan2016developing}, it builds a regrasp graph by connecting nodes between
given initial and goal placements for an object, and searches the graph to
find a sequence of pick-and-place sub-tasks. This generates a sequence of
states, grasps, transit and transfer motion, connecting the starting and ending
nodes. This paper improves the previous planner by evaluating if the state of
the robot-object system is stable, in order to preserve its balance during
searching. If the state is IK-unfeasible, generates a collision, or makes the
robot fall out of balance, the corresponding node is removed and the search continues until a set of nodes
connecting the initial and goal poses is found.

The stability of a given state is computed using the CoM of the robot-object
system and the robot support polygon, this provides a means to measure how
stable the robot-object system is. By determining the CoM of the system and its
projection over the robot support polygon plane, we can determine if the robot
is stable \cite{mcghee1968stability}. For this work, it is assumed that the
robot is standing on an even and flat surface while performing regrasp tasks.
This enables the planner to evaluate the states of the robot without
extra sensors and complicated calculations.

\section{Stability Measurement and Constraints}

In order to generate stable motion sequences for the robot, the planner
determines the global position of the robot-object system CoM for each robot
state. Given the CoM global position and the robot feet position, the planner
checks if the projection of the CoM is inside the support polygon of the robot
and its distance to the polygon edges. If the distance is below a
given threshold, the state is discarded and a new graph search is started.

The proposed method can also be used to explore the stability of a task: By
randomly checking the stability of the possible robot states for the task
completion in a biased fashion, a highly task-related stability measurement can
be performed by determining the ratio of stable evaluated robot states and the
total amount of evaluated poses.

\subsection{Calculation of robot-object system CoM}
In our simulator, a local coordinate system is defined and used to determine
the changes of joints. With the rotation matrices of
the robot links, it is possible to obtain the global position of the robot limbs
and the manipulated object, as well as their CoMs during regrasp. Thus,
we compute the CoM of the whole system, $\Gamma^{S}(x,y,z)$ using
\begin{equation} \label{eq:system CoM position}  
\Gamma^{S}(x,y,z)=
\frac{M\Gamma^{robot}(x,y,z)+m\Gamma^{object}(x,y,z)}{M+m}
\end{equation}
where $M$ and $m$ are the masses of the robot and the object respectively and
$\Gamma^{robot}(x,y,z)$ and $\Gamma^{object}(x,y,z)$ are their CoMs.

Once the CoM of the system is calculated, the robot support polygon is computed
using the convex hull formed by its feet. For a robot pose to be considered
stable, the projection of $\Gamma^{S}(x,y,z)$ must be inside the support polygon
and the minimum distance between the projection and the edges of the convex hull
polygon must be higher than a given threshold. If a robot pose/state does not
comply with these constraints, the pose is discarded and a new search is
started.

\subsection{Task stability measurement}

Since humanoid robots can perform one-hand or dual-arm regrasp, it is
of interest to find task-related stability estimations for the different
regrasping solutions a humanoid robot may have for a task. The state evaluation
is performed using a energy based search method inspired by
RRT, which presents a biased search towards low-cost regions such that solution
paths remain close to minimal work paths \cite{jaillet2008transition}.

The biased search can be used to perform a stability measurement highly related
to a specific task: by performing a continuous state exploration, several random
but goal-biased robot poses can be evaluated and a ratio of stable poses over
explored poses can be calculated by using Eqn.\eqref{eq:task stability
ratio}.
\begin{equation} \label{eq:task stability ratio} R_{hand}=
\frac{C_{hand}}{C_{hand}+U_{hand}}\end{equation}
where $R_{hand}$ is the
stability ratio between the stable poses $C_{hand}$ and the total amount of
explored poses which also includes the unstable poses $U_{hand}$. This ratio is
calculated for a particular robot leg stance and a given hand configuration
$hand$, the hand configuration indicates which hand is going to be used for the
task in the case of a one-armed regrasp task or in which order the hands will be
used, in case of dual-arm regrasp. Given the randomness of the RRT exploration
and the introduced bias towards the starting and ending points of a given path
for performing the task, this method represents a task-related stability
estimation, which can be used to choose between single-arm and dual-arm regrasp
or between the use of the left or right hand for a given regrasp task,
maximizing the robustness and stability of the motions of the robot by using the
hand configuration with the highest stability ratio $R_{hand}$.

\section{Simulations and Real-world Results} 

To test the constrained planner, a series of simulations were performed using an
HRP5P model. In the simulations, the robot is required to change the pose of a
given object and move it to a goal position, as seen in Fig.\ref{fig:HRP5P SIM
comparison thresholds}. The HRP5 has 35 DoFs making it a suitable platform for
highly complex motions and dexterous manipulation. Four different stances were
simulated, in each case, the task stability measurement method was used to
determine the stability ratio for the different hand configurations and stances
available to the robot. The planner is also tested using a real robot. 
Two challenging tasks involving object regrasping were
given to the robot to test the planner ability to generate a motion path for the
robot that complies with the imposed constraints to its CoM location. The
results of these tests are discussed in this section, too.

\begin{figure*}[!htbp]
\centering
 \includegraphics[width=.9\textwidth]{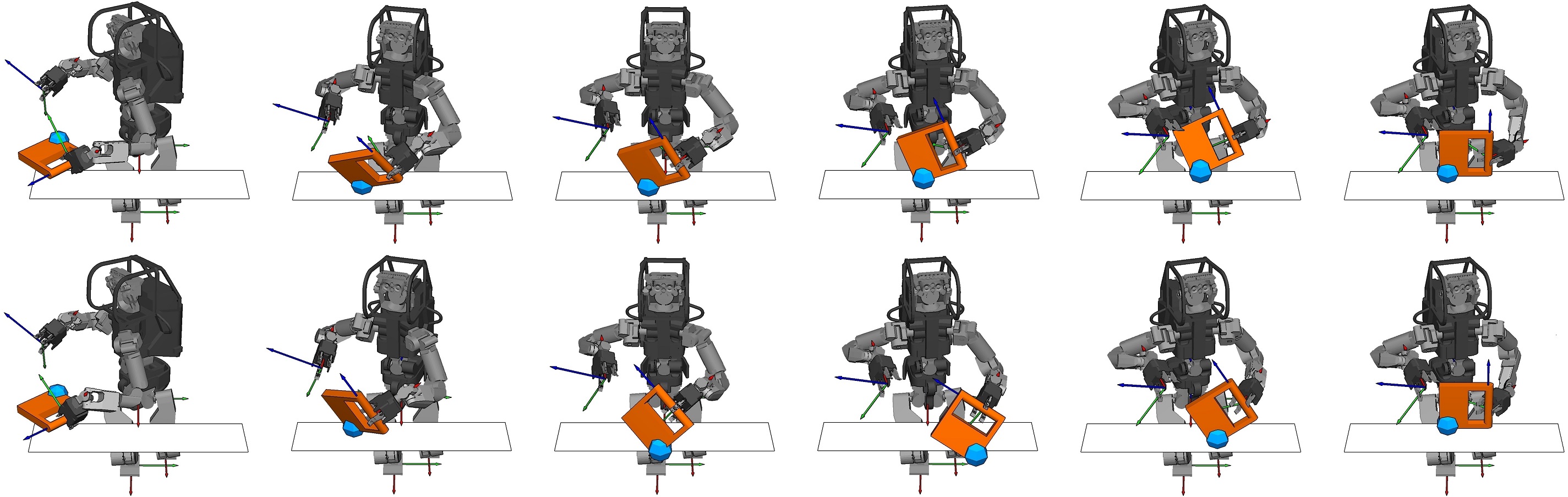}
 \caption{Comparison between planned motion with different thresholds. In
 the upper case, the planner is given a 30 $mm$ threshold whilst in
 the lower case the threshold is 0 $mm$. The difference in constraints
 causes the planner to generate an alternative motion, preventing the
 object CoM, represented by the blue sphere, from traveling too far from the
 support polygon when the threshold is higher.}
 \label{fig:HRP5P SIM comparison thresholds}
 \end{figure*}

\subsection{Simulations}

\subsubsection{Upright stance}
\begin{figure*}[!htbp]
\centering
 \includegraphics[width=.9\textwidth]{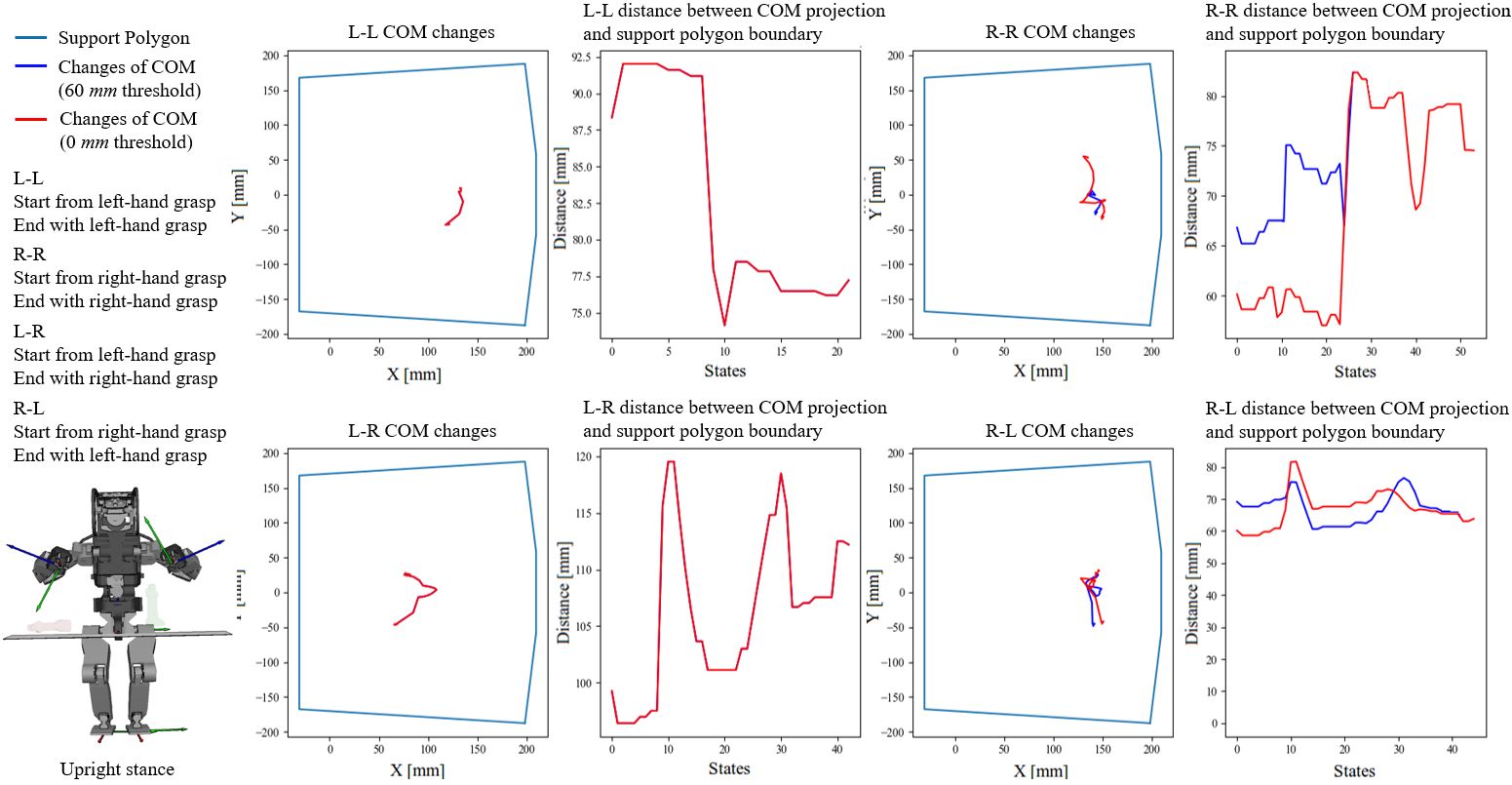}
 \caption{CoM changes for the robot-object system while performing regrasp
 tasks with an upright stance and different hand configurations and thresholds.
 The results show the different CoM changes generated by the planner to maintain
 the stability of the task over two given thresholds. NOTE: When the results 
 are the same for 0 mm and 60 mm thersholds, the blue curves (60 mm) are hidden
 by the red curves (0 mm). All plots in following figures are the same.}
 \label{fig:HRP5P upright stance com movement and distance comparison}
 \end{figure*}

In this case, the robot assumes an upright posture, with its feet in
symmetrical positions. A minimum stability threshold of 60 $mm$ was given to the
regrasp planner. The manipulated object has a mass of 8 $Kg$. Its CoM
is located at (150, 80, 150) $mm$ in the object's local
coordinate system.
The object has its initial position on the robot right-hand side and it is
laying sideways on a table, while the object's goal position is on the robot's
left-hand side. The results of task stability measurement are shown in Table
\ref{Tab: Upright Stance: Path evaluations }. The changes of CoM in the
projection plane are shown by blue curves in Fig.\ref{fig:HRP5P upright stance com movement and distance comparison}. In the
same figure, a comparison to threshold 0 $mm$ is shown in red curves.
              
\begin{table}[h]
\centering
\caption{Upright Stance: Path evaluations}
\label{Tab: Upright Stance: Path  evaluations }
\resizebox{\columnwidth}{!}{%
\begin{threeparttable}
\begin{tabular}{lccc}
\toprule
Hand configuration            & $C_{hand}$ & $U_{hand}$ & $R_{hand}$ \\
\midrule
Start from left hand - End with left hand      & 10529      &
15 & 0.9986
\\
Start from right hand - End with right hand       & 15991      & 1801       &
0.8998
\\
Start from left hand - End with right hand  & 11189      & 41         & 0.9963    
\\
Start from right hand - End with left hand   & 5391       & 7351       & 0.4231   
\\
\bottomrule
\end{tabular}
\begin{tablenotes}
\item[Meanings of abbreviations] $C_{hand}$: \# of stable poses; $U_{hand}$: \#
of unstable poses; $R_{hand}$: Ratio between $C_{hand}$ and
$C_{hand}$+$U_{hand}$; Hand configuration: The robot will plan a
motion between two left-hand, right-hand, or left-to-right-hand grasps. If the
grasps were from different hands, the robot will plan a handover motion.
\end{tablenotes}
\end{threeparttable}
}
\end{table}

\subsubsection{Staggered stance} 
\begin{figure*}[!htbp]
\centering
 \includegraphics[width=.9\textwidth]{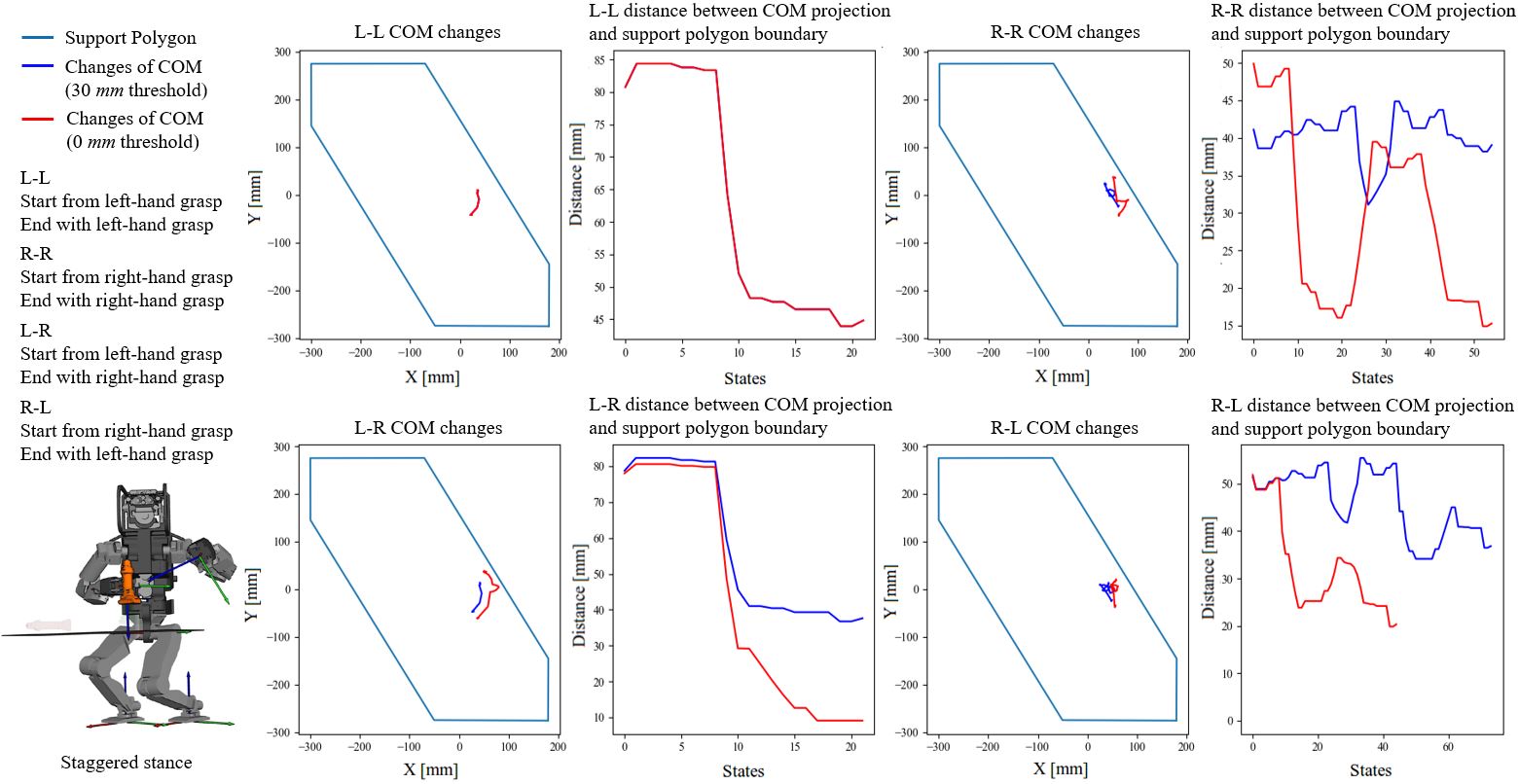}
 \caption{CoM changes of the system while performing regrasp
 with a staggered stance and different hand configurations and thresholds.}
 \label{fig:Staggeredstancecommovementupdate}
\end{figure*}

For this simulations, the robot assumes a staggered posture with its right leg
50 $mm$ in front of the robot local coordinate system and its left foot 200
$mm$ behind. In this case, the object has a mass of 5 $Kg$ and the minimum
threshold given to the planner was of 30 $mm$. The object has the same origin
and goal poses as the upright stance simulation, but its CoM is located at
(-150, 80, 150) $mm$ of the object's local coordinate system. The results of
task stability measurement are shown in Table \ref{Tab: Staggered Stance:
Path evaluations }. The changes of CoM in the projection plane are shown in
blue curves in Fig.\ref{fig:Staggeredstancecommovementupdate}. A comparison with
the minimum threshold set to 0 $mm$ is shown in red curves in the same figure.

\begin{table}[h]
\centering
	\caption{Staggered Stance: Path evaluations}
	\label{Tab: Staggered  Stance: Path evaluations }
	\resizebox{\columnwidth}{!}{%
	\begin{tabular}{lccc}
	\toprule
	Hand configuration & $C_{hand}$ & $U_{hand}$ & $R_{hand}$ \\
	\midrule
	Start from left hand - End with left hand & 6447 & 135 & 0.9795 \\
	Start from right hand - End with right hand & 250 & 90 & 0.64        \\
	Start from left hand - End with right hand & 2135 & 6482 & 0.2478      \\
	Start from right hand - End with left hand & 1370 & 390 & 0.7784 \\
	\bottomrule
	\end{tabular}
	}
\end{table}

\subsubsection{Crouched stance} 
In this case, a crouching posture of the robot is used for simulation.
The mass of the object is set to 15 $Kg$ and the distance
threshold between the CoM of the robot-object system and the support polygon is
set to 60 $mm$. The object CoM is located at (-150, 80, 150) $mm$ in the object
local coordinate system. The recorded data from the task stability measurement
is shown in Table \ref{Tab: Crouched  Stance: Path  evaluations }. Comparison of
CoM changes with threshold 60 $mm$ and 0 $mm$ is shown in
Fig.\ref{fig:crouchedstancecommovementupdate}.

\begin{table}[!htbp]
\centering
	\caption{Crouched Stance: Path evaluations}
	\label{Tab: Crouched  Stance: Path  evaluations }
	\resizebox{\columnwidth}{!}{%
	\begin{tabular}{lccc}
	\toprule
	Hand configuration & $C_{hand}$ & $U_{hand}$ & $R_{hand}$ \\
	\midrule
	Start from left hand - End with left hand & 12242 & 0 & 1  \\
	Start from right hand - End with right hand & 654 & 0 & 1 \\
	Start from left hand - End with right hand & 6901 & 0 & 1 \\
	Start from right hand - End with left hand & 14754 & 0 & 1 \\
	\bottomrule
	\end{tabular}
	}
\end{table}

\subsubsection{One legged stance} 
For the final set of simulations, the robot was given a highly unstable posture,
a one-legged stance with the right foot set directly under the robot CoM and the
left foot raised 50 $mm$ above the ground. The robot is required to perform a
regrasp task with a bar-like object with 5 $Kg$ of mass. The object CoM is located at
coordinates (0,0,200) $mm$ and the stability threshold for the task is set to
30 $mm$. The data gathered from this experiment is shown in table
\ref{Tab: One-Legged Stance: Path  evaluations }. The comparison with threshold
0 $mm$ is shown in Fig.\ref{fig:Oneleggedstancecommovementupdate}.

\begin{table}[!htbp]
\centering
	\caption{One-Legged Stance: Path evaluations}
	\label{Tab: One-Legged Stance: Path  evaluations }
	\resizebox{\columnwidth}{!}{%
	\begin{tabular}{lccc}
	\toprule
	Hand configuration & $C_{hand}$ & $U_{hand}$ & $R_{hand}$ \\
	\midrule
	Start from left hand - End with left hand & 2628 & 0 & 1 \\
	Start from right hand - End with right hand & 4299 & 2401 & 0.6416 \\
	Start from left hand - End with right hand & 2558 & 2444 & 0.5113 \\
	Start from right hand - End with left hand & 1306 & 88 & 0.9368 \\
	\bottomrule
	\end{tabular}
	}
\end{table}

\subsection{Analysis of the simulation results }
For the upright stance, the planner did not need to remove unstable poses from 
the regrasp graph when the robot used a left-handed one-armed configuration, or
when it used a Left-Right dual-armed configuration, making the planned motions
for the different thresholds equal as seen in Fig.\ref{fig:HRP5P upright stance
com movement and distance comparison}. On the other hand, some path corrections
were performed by the planner for the right hand configuration and left-right
hand configuration. The results show that the system CoM tends to move more
towards the support polygon edge when the robot finishes its task using the
right hand, but the planner is able to generate a path that keeps the CoM
distance above the given threshold. These results also agree with the data shown
in Table \ref{Tab: Upright Stance: Path  evaluations } which illustrates the
left and left-right hand configurations as the most stable for the task.

For the staggered stance, the robot showed higher stability while
performing the regrasp task using only its left arm, with 0 unstable poses
recorded while the robot executed the task and a task evaluation of 0.9795 as
seen in Table \ref{Tab: Staggered  Stance: Path  evaluations } and
Fig.\ref{fig:Staggeredstancecommovementupdate}. Meanwhile, the most unstable
hand configurations involve right hand being used to put the object in its final
position, this shows that the twisting motion performed by the robot to lower
down the object to its goal position with its right hand introduces undesired
instability for this particular task. The staggered position changes the robot
support polygon making the robot balance more susceptible to sideways motions.
Since the right foot did not change positions, compared to the previous
experiment, the stability of action of picking up the object, which is located
on the right-hand side of the robot, does not change noticeably. On the other
hand, since the left leg does not provide as good a support compared to
the previous experiments, sideways motions to the left-hand side of the robot
introduce instability to the robot.

\begin{figure*}[!htbp]
\centering
 \includegraphics[width=.9\textwidth]{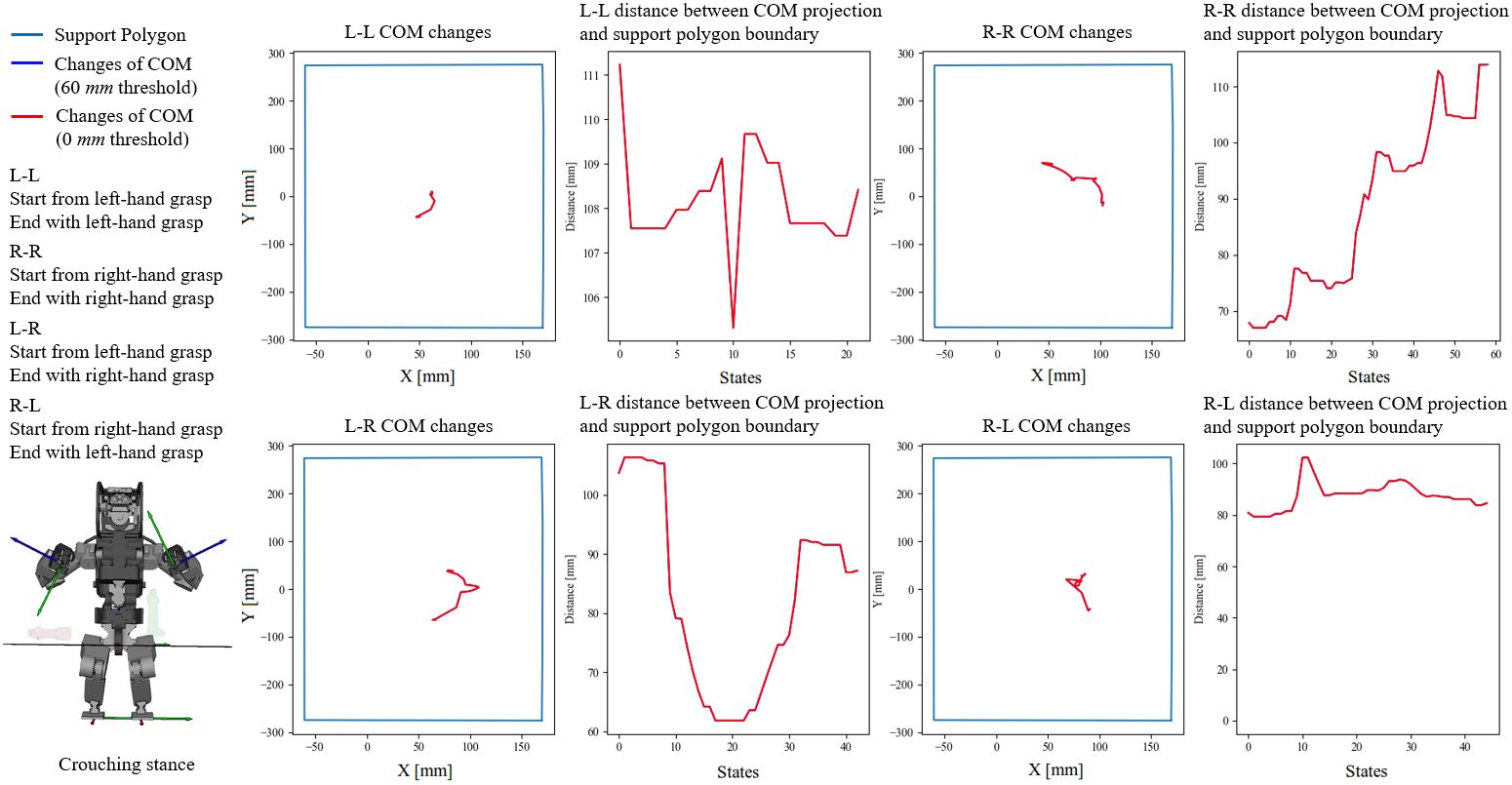}
 \caption{CoM changes of the system while performing regrasp
 with a crouched stance and different hand configurations and
 thresholds.}
 \label{fig:crouchedstancecommovementupdate}
\end{figure*}

The crouched stance aligns the robot CoM with its
support polygon and also lowers its height. The posture puts the robot in a more
stable state, which allowed us to increase the weight of the object to 8 Kg
and the threshold to 60 mm while still guaranteeing stable states for planning. With a
stability ratio of 1 and 0 unstable poses encountered as seen in Table \ref{Tab:
Staggered  Stance: Path evaluations } and
Fig.\ref{fig:crouchedstancecommovementupdate}, we were able to find a stance
that maximizes the stability of the task, making the task stability measurement
a useful tool for finding not only the most stable hand configuration, but also to
compare the stability of different stances for the same task.

\begin{figure*}[!htbp]
\centering
 \includegraphics[width=.9\textwidth]{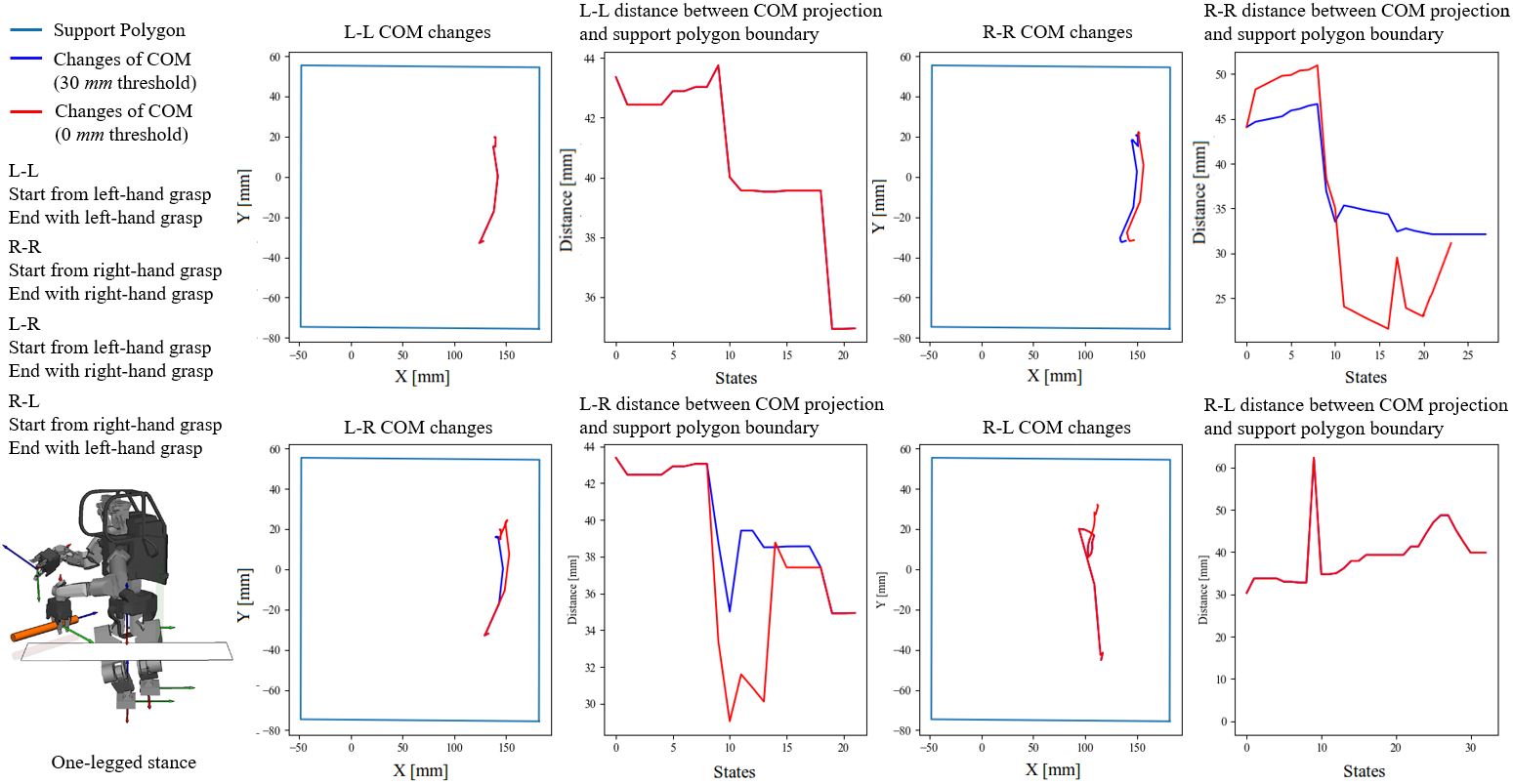}
 \caption{CoM changes of the system while performing regrasp
 with a one-legged stance and different hand configurations and
 thresholds.}
 \label{fig:Oneleggedstancecommovementupdate}
 \end{figure*}

The staggered stance is challenging. The
results indicate that the left-hand configuration is the most stable choice for
the robot, with a stability ratio of 1, as shown in table \ref{Tab: One-Legged
Stance: Path  evaluations }. These results show that twisting motion performed
by the use of the right hand to put the object in its goal pose introduces
instability to the robot-object system, as seen in
Fig.\ref{fig:Oneleggedstancecommovementupdate}. Since the left leg is not
supporting the body and it does not move with the rotation of the waist, there
is more unsupported mass on the left-hand side. When the robot twists
its waist to place the object with the right hand, the CoM shifts further to the
left, making the task more unstable. Note that
the constraints for the one-legged stance are softened. The 
stability ratio is therefore sometimes higher than other stances.

In several cases, the comparison between CoM changes with a threshold
of 0 $mm$ and other threshold values results in a noticeable difference in the
robot motions, as seen in Fig.\ref{fig:HRP5P
SIM comparison thresholds}. On the other hand, when the conditions of the
robot-object system allow the robot to perform the task without incurring in
unstable poses for a given threshold, the planner generates the same motion
for lower threshold values, as seen in Fig.\ref{fig:crouchedstancecommovementupdate}. Since the planner does not
necessarily generate the most stable path but assures that it maintains a minimum
of stability, these results are expected. Meanwhile, the planning time is 
still comparable to \cite{wan2016developing}.

\begin{figure*}[!htbp]
        \centering
        \begin{subfigure}{\textwidth}
                \centering
                \includegraphics[width=.9\textwidth]{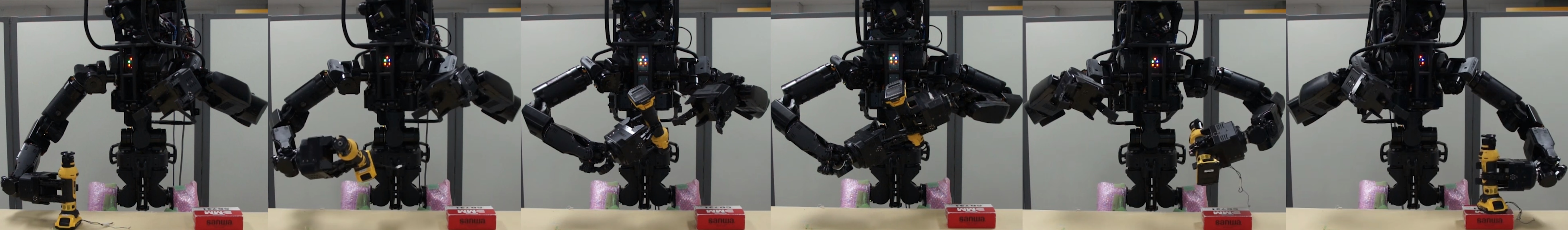}
                \caption{The HRP5P performing a dual-arm regrasp task.
                The robot is commanded to move the drill using a
                right-to-left-hand regrasp.}
        \vspace{0.05in}
        \end{subfigure}
        ~
        \begin{subfigure}{\textwidth}
                \centering
                \includegraphics[width=.9\textwidth]{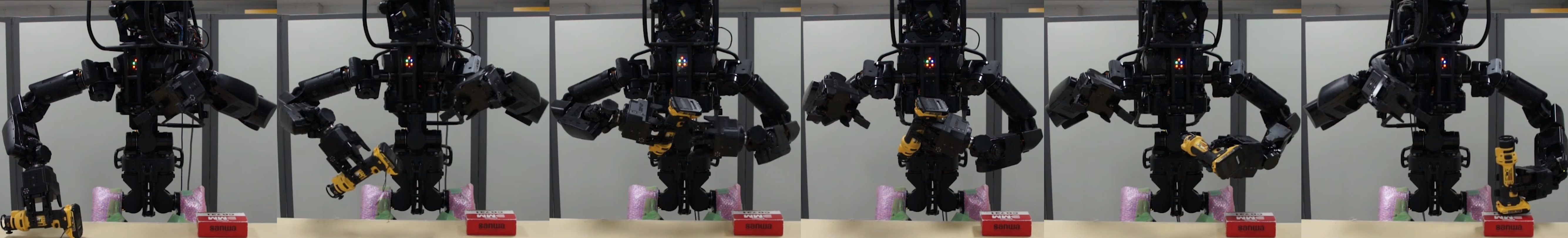}
                \caption{The second dual-arm regrasp task.
                In this case, the object is laying sideways initially. The goal
                is in an upright pose.}
        \end{subfigure}
        \caption{Motion sequences of the HRP5P robot reorienting an
        electric drill using dual-arm regrasp.}
        \label{fig:hrp5preal}
\end{figure*}

\subsection{Real-world experiments}

Following the results obtained in the simulations, we implemented our
CoM-restricted regrasp planner in the HRP5P humanoid robot platform. The HRP5P
represents the 5th generation of the Humanoid Robotics Project \cite{inouehrp}.
In this case, the planning of the robot motions was performed off-line with our
simulator. The motion sequences generated for this tasks were then given to the
robot to perform. The setting and environment of the robot are composed by a
table, an electric drill and a small box as seen in Fig.\ref{fig:hrp5preal}(a).

To test the planner, the robot was given two dual-arm regrasping tasks.
In the first task, the robot is required to move the drill by picking it up
using a right-hand grasp and placing it down using a left-hand grasp. The
planner is able to generate a motion sequence for the robot while keeping the
robot-object system CoM above a minimum threshold distance to the support
polygon edges of 55 $mm$. The execution of the planned result is
shown in Fig.\ref{fig:hrp5preal}(a).
In the second task, the object is laying sideways initially and the goal is to
reorient it to an upright pose. The robot successfully completed the task
without tipping over or colliding with its environment or itself. The execution
is shown in Fig.\ref{fig:hrp5preal}(b). Details are in the supplementary video.

\section{Conclusions}
This paper presented a CoM based planner that planned regrasp motions while
preserving robot balance.
Simulation results showed that the planner eliminates several robot states
that could be too unstable for the robot, generating different motions that keep
the system CoM distance to the support polygon edges above the given threshold
without noticeably increasing computation time. The results also revealed that the proposed task stability
measurement could be used to evaluate the stability of a task with uncommon, or
sub-optimal stances which a robot might be forced to take, for reasons such as
obstacles or damaged parts. The simulation results were executed by an HRP5P
robot, demonstrating its practicality.


\bibliographystyle{IEEEtran}
\balance
\bibliography{paperDaniel}

\end{document}